\pdfoutput=1

\documentclass[11pt,table]{article}

\usepackage[]{acl}

\usepackage{times}
\usepackage{latexsym}
\usepackage{graphicx}
\usepackage{multicol}
\usepackage{multirow}
\usepackage{url}
\usepackage{hyperref}
\usepackage{booktabs}

\usepackage{tikz}
\usetikzlibrary{trees}

\usepackage[T1]{fontenc}

\usepackage[utf8]{inputenc}

\usepackage{microtype}

\usepackage{inconsolata}

\usepackage{amssymb}
\usepackage{enumitem}
\usepackage{amsmath}

\definecolor{Gray}{gray}{0.93}
\usepackage{colortbl}

\newcommand{\inputPose}{X}
\newcommand{\reconPose}{\hat{X}}
\newcommand{\latentContinuous}{Z_e}
\newcommand{\latentQuantized}{Z_q}
\newcommand{\codebook}{C}
\newcommand{\codebookSize}{K}
\newcommand{\latentDim}{L_c}
\newcommand{\numLatents}{N_p}
\newcommand{\numFrames}{T}
\newcommand{\poseDim}{D}

\newcommand{\lossBaseline}{\mathcal{L}_{\text{Baseline}}}
\newcommand{\lossRecon}{\mathcal{L}_{\text{recon}}}
\newcommand{\lossCodebook}{\mathcal{L}_{\text{codebook}}}
\newcommand{\lossCommit}{\mathcal{L}_{\text{commit}}}

\newcommand{\betaCommit}{\beta}

\newcommand{\gammaDiversity}{\gamma}

\newcommand{\inputLH}{\inputPose^{(\textsc{lh})}}
\newcommand{\inputRH}{\inputPose^{(\textsc{rh})}}
\newcommand{\inputNMM}{\inputPose^{(\textsc{nmm})}}
\newcommand{\inputMOVL}{\inputPose^{(\textsc{movl})}}
\newcommand{\inputMOVR}{\inputPose^{(\textsc{movr})}}
\newcommand{\inputLOC}{\inputPose^{(\textsc{body})}}

\newcommand{\modelBase}{\text{Baseline}}
\newcommand{\modelPD}{\text{VQ-ASL-PD}}
\newcommand{\modelPSS}{\text{VQ-ASL-PSS}}
\newcommand{\modelFull}{\text{VQ-ASL (Full)}}

\newcommand{\model}{VQ-ASL}
\newcommand{\modelshort}{VQ-ASL}

\title{Phonological Representation Learning for Isolated Signs \\ Improves Out-of-Vocabulary Generalization }

\author{
Lee Kezar \\
University of \\ Southern California \\
\texttt{lkezar@usc.edu} \\
\And
Zed Sehyr \\
Chapman \\ University \\
\texttt{sehyr@chapman.edu} \\
\And
Jesse Thomason \\
University of \\ Southern California \\
\texttt{jessetho@usc.edu} \\
}

\begin{document}
\maketitle
\begin{abstract}
Sign language datasets are often not representative in terms of vocabulary, underscoring the need for models that generalize to unseen signs.
Vector quantization is a promising approach for learning discrete, token-like representations, but it has not been evaluated whether the learned units capture spurious correlations that hinder out-of-vocabulary performance.
This work investigates two phonological inductive biases: Parameter Disentanglement, an architectural bias, and Phonological Semi-Supervision, a regularization technique, to improve isolated sign recognition of known signs and reconstruction quality of unseen signs with a vector-quantized autoencoder.
The primary finding is that the learned representations from the proposed model are more effective for one-shot reconstruction of unseen signs and more discriminative for sign identification compared to a controlled baseline.
This work provides a quantitative analysis of how explicit, linguistically-motivated biases can improve the generalization of learned representations of sign language.
\end{abstract}

\begin{figure}
    \centering
    \includegraphics[width=0.5\textwidth]{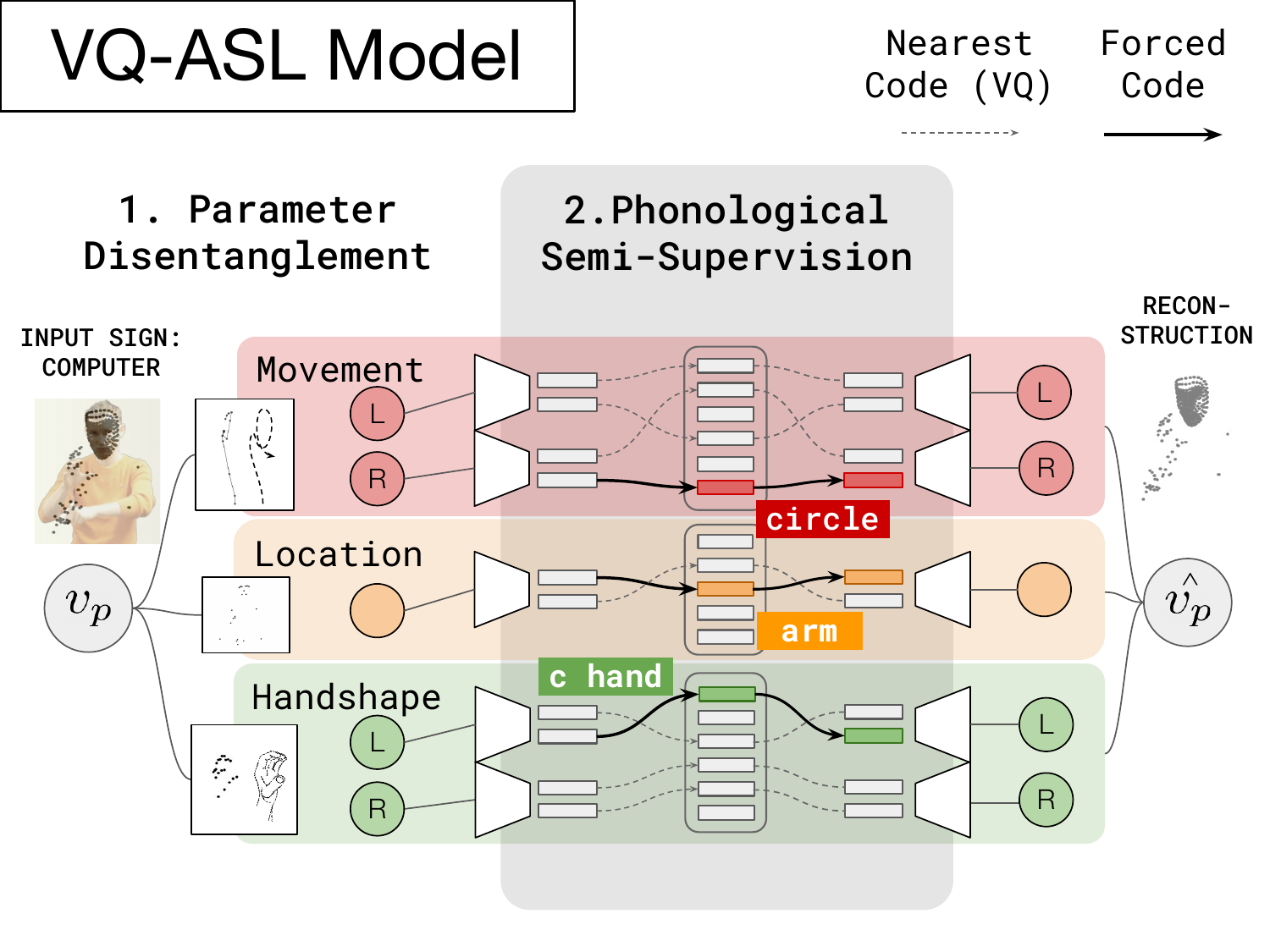}
    \caption{We show that out-of-vocabulary signs are more accurately reconstructed when trained under two phonological inductive biases: first, \textit{disentangling} the input stream by ASL parameter; then, encoding each stream as a sequence of learned components, some of which are aligned with expert labels.}
    \label{fig:teaser}
\end{figure}

\section{Introduction} \label{sec:intro}
The development of robust sign language models is often constrained by the limited scale and vocabulary of available datasets \citep{bragg2019interdisciplinary, desisto2022challenges}.
This data scarcity makes the ability to generalize to out-of-vocabulary (OOV) signs a significant challenge for the field.
Sign languages are highly productive systems, where a finite set of phonological features, such as handshape, location, and movement patterns can be combined to form a vast lexicon \citep{stokoe1960structure, brentari1998prosodic}.
A model that learns to represent these underlying phonological components of signs can improve sign recognition~\citep{kezar2023improving}, and we hypothesize such phonological learning bias could improve reconstruction of novel combinations of those components in unseen signs.

Vector-Quantized Variational Autoencoders (VQ-VAEs) have emerged as a powerful method for learning discrete latent representations of data, which are attractive for their potential use as tokens in sequence models \citep{oord2017neural, razavi2019generating}.
Prior work has explored VQ for creating data-driven representations of signs, often as an alternative to linguistic glosses \citep{abzaliev2024unsupervised, tasyurek2025disentangle}.
However, standard VQ models are trained with a compression objective that may encourage the learning of "tangled" representations, where the learned codes capture spurious, dataset-specific correlations that do not generalize to unseen signs~\citep{higgins2017beta}.
This reflects a theoretical limitation formally proven by \citet{locatello2019challenging}, who showed that the unsupervised learning of disentangled representations is "fundamentally impossible" without inductive biases on both the models and the data.
This finding shows that a search for principled sources for such biases is not just motivated, but necessary.

This paper therefore investigates whether phonologically-motivated inductive biases can improve a VQ-VAE's ability to reconstruct and recognize out-of-vocabulary (OOV) signs.
We draw these biases from the Prosodic Model \citep{brentari1998prosodic} and the ASL-LEX 2.0 database \citep{sehyr2021asllex}, two lexicon-wide descriptions of ASL structure, and implement them through two mechanisms (Figure \ref{fig:teaser}).
The first is \textit{Parameter Disentanglement} (PD), an architectural method that uses a multi-stream VQ-VAE to learn separate codebooks for distinct articulators and movement parameters.
The second is P\textit{honological Semi-Supervision} (PSS), a supervisory method that uses an auxiliary classification loss with expert phonological labels to regularize the latent codebooks and align them with established linguistic features.

This paper proceeds as follows: Section \ref{sec:background} reviews relevant background literature.
Section \ref{sec:framework} details the VQ-ASL framework.
Section \ref{sec:experiments} describes the experimental setup for a controlled ablation study.
Section \ref{sec:results} presents the results, and Section \ref{sec:discussion} discusses their implications and the limitations of the study.
The main finding is that the proposed interventions offer complementary benefits, improving OOV generalization for both sign reconstruction and recognition, and revealing a trade-off between reconstruction fidelity and the discriminative quality of the learned representations.

\section{Background} \label{sec:background}
We review prior work in sign language representation learning, the theoretical foundations of disentangled representation learning, and the linguistic models that motivate our proposed inductive biases.

\subsection{Representation Learning for Sign Language}
\label{subsec:repr_learning_sl}
The popular input modality for modern sign language recognition is skeletal pose data, extracted from video using tools like MediaPipe or OpenPose \citep{lugaresi2019mediapipe, cao2019openpose}.
This representation can be seen as a form of \textit{phonetic bias}, because it abstracts away from signer-specific visual details like clothing or background, focusing instead on the underlying articulatory movements while enhancing privacy \citep{bragg2020exploring}.
Various neural architectures have been employed to learn features from this data.
3D Convolutional Neural Networks (3D CNNs) can capture spatio-temporal features directly from video, but are computationally intensive and may learn spurious visual cues \citep{pu2021learning}.
Graph Convolutional Networks (GCNs) are naturally suited to skeletal data, as they explicitly model the topological structure of the human body and can learn the dynamic relationships between joints over time \citep{yan2018spatial,sl-gcn}.
More recently, contrastive learning objectives have been used to align visual representations of signs with textual descriptions, improving the grounding of the learned features without supervision \citep{jiang-etal-2024-signclip, hao2021selfsupervised}.

The goal of obtaining discrete representations for signs includes manual and learned efforts.
Formal symbolic systems like SignWriting provide a manual transcription system, analogous to written text \citep{sutton1990signwriting}.
Data-driven approaches, particularly those using vector quantization, have also been explored to learn discrete tokenizations of sign language, often as an intermediate step for sign language translation or production \citep{moryossef2021unsupervised, saunders2020everybody}.
A recent preprint, "Disentangle and Regularize," also investigates articulator-based disentanglement for sign language production, though with a focus on continuous latent spaces and different regularization techniques \citep{tasyurek2025disentangle}.
Our work is distinct in its focus on vector quantization and the use of phonological semi-supervision to structure the discrete codebooks for OOV generalization in isolated sign recognition.

\subsection{Disentangled Representation Learning}
\label{subsec:disentangled_learning}
The goal of disentangled representation learning is to produce a latent space where each dimension, or group of dimensions, corresponds to a distinct, meaningful factor of variation in the data \citep{higgins2017beta}.
A model that successfully disentangles the underlying generative factors is hypothesized to exhibit better compositional generalization, data efficiency, and robustness.
In the context of signing, these factors may be defined as the five phonological parameters: handshape, palm orientation, location, movement, and non-manual markers.

A foundational work in this area is the $\beta$-VAE, proposed by \citet{higgins2017beta}.
This framework modifies the standard Variational Autoencoder (VAE) objective by introducing a hyperparameter, $\beta$, that increases the weight of the Kullback-Leibler (KL) divergence term in the loss function.
A value of $\beta > 1$ imposes a stronger constraint on the information capacity of the latent bottleneck, forcing the model to learn a more efficient, and therefore more disentangled, representation by balancing reconstruction accuracy against the statistical independence of the latent variables.

However, the pursuit of unsupervised disentanglement faced a significant theoretical challenge from \citet{locatello2019challenging}.
They provided a formal proof demonstrating that for any dataset generated from disentangled latent factors, there exists an infinite family of transformations that can produce a perfectly entangled latent space while yielding the exact same data distribution.
As an unsupervised model only has access to the observed data, it cannot distinguish between the true disentangled model and its entangled counterparts.
Their conclusion is sobering: "the unsupervised learning of disentangled representations is fundamentally impossible without inductive biases on both the models and the data."
This study provides the central motivation for this work, which seeks to identify and implement a principled source for such biases.

\begin{figure}
    \centering
    \begin{tikzpicture}[
      level 1/.style={sibling distance=30mm, level distance=12mm},
      level 2/.style={sibling distance=20mm},
      level 3/.style={sibling distance=20mm},
      level 4/.style={sibling distance=17mm},
      every node/.style={align=center},
      leaf/.style={draw, rounded corners=2pt, inner sep=2pt}
      ]
    
    \node {Root node}
      child {node {Inherent}
        child {node {Articulator}
          child {node[leaf] {Non-manual\\articulators}}
          child {node {Manual\\articulators}
            child {node[leaf] {Hand$_2$\\(non-dom.)}}
            child {node[leaf] {Hand$_1$\\(dom.)}}
          }
        }
        child {node[leaf] {Place of\\articulation}}
      }
      child {node {Prosodic}
        child {node {Setting\\change}
          child {node[leaf] {Path}}
        }
      };
    
    \end{tikzpicture}

    \caption{Brentari's Prosodic Model \citep{brentari1998prosodic} is hierarchical; the boxed variables represent disentangled parameters in this work.}
    \label{fig:brentari}
\end{figure}
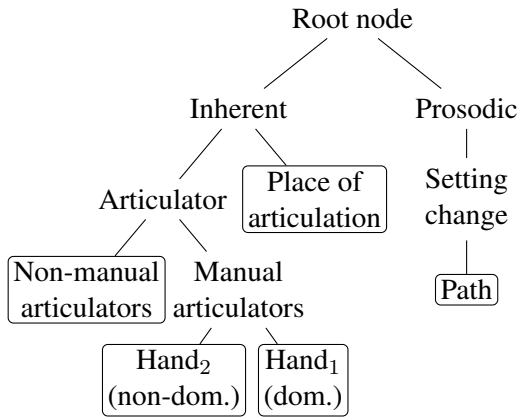

\subsection{Linguistic Theory as a Stratified Inductive Bias}
\label{subsec:linguistic_bias}
Our work responds to the challenge posed by \citet{locatello2019challenging} by using formal linguistic theory as a robust source of inductive biases, an approach aligned with Knowledge-Infused Learning (KiL) \citep{gaur2022advancing}.
We apply this knowledge in a stratified manner for a deep infusion: a low-level architectural bias enforces parameter disentanglement on the articulators, while a higher-level phonological semi-supervision acts as a conceptual bottleneck that guides the model through linguistically-defined concepts \citep{koh2020concept}.
The specific knowledge for this work is the Prosodic Model of sign language phonology, developed by \citet{brentari1998prosodic}.

Brentari's model describes signs as being composed of a hierarchical and simultaneous arrangement of phonological parameters (Figure \ref{fig:brentari}).
This linguistic framework provides a blueprint for a \textit{stratified} inductive bias, which we implement using mechanisms from Knowledge-Infused Learning \citep{gaur2022advancing}.
The architectural bias in our model is informed by Brentari's factorization of signs into articulators (hands, face), place of articulation (location), and prosodic features (movement).
We additionally use the discrete, contrastive features described by the theory, such as the handshapes cataloged in ASL-LEX 2.0 \citep{sehyr2021asllex}, to regularize the learning process through an auxiliary loss.
For a complete description of the ASL-LEX features, see Appendix \ref{app:feature_assignment}.
Despite the availability of several phonological models (e.g., H-M-H, Liddell \& Johnson, 1989; Dependency Model, van der Kooij, 2002), we adopt Brentari's prosodic model as best suited for semi-supervised learning.
It offers (1) a clear inventory of contrastive features aligned with linguistic theory, and (2) a hierarchical structure that maps well onto the discrete components produced by our model. 

By leveraging both architectural and regularization-based KiL strategies, guided by different strata of Brentari's Prosodic Model, we aim to provide the structured priors necessary for learning a meaningful and generalizable disentangled representation of signs.

\section{The VQ-ASL Framework} \label{sec:framework}
This section provides a detailed technical description of the proposed VQ-ASL framework.
It begins by detailing the baseline architecture, a Transformer-based VQ-VAE (Section \ref{subsec:baseline_arch}), and subsequently elaborates on the two novel, phonologically-motivated interventions: Parameter Disentanglement (PD) and Phonological Semi-Supervision (PSS) (Section \ref{subsec:phon_biases}).

\subsection{Baseline Transformer VQ-VAE}
\label{subsec:baseline_arch}
The baseline model is a Vector-Quantized Variational Autoencoder (VQ-VAE), which learns a discrete latent representation of an input sequence~\citep{oord2017neural}.
The encoder and decoder components are implemented using Transformer architectures~\cite{vaswani2017attention} to effectively model the sequential nature of sign language pose data.

The \textbf{encoder}, $E$, takes a sequence of pose vectors $\inputPose \in \mathbb{R}^{\numFrames \times \poseDim}$ as input, where $\numFrames$ is the number of frames and $\poseDim$ is the dimensionality of the pose vector for each frame.
The Transformer encoder processes this sequence and outputs a set of $\numLatents$ continuous latent vectors, $\latentContinuous = E(\inputPose)$, where $\latentContinuous \in \mathbb{R}^{\numLatents \times \latentDim}$ and $\latentDim$ is the latent channel dimension.
In the baseline model, these $\numLatents$ vectors form a single, unstructured latent sequence.

The \textbf{vector quantization layer} serves as the information bottleneck.
It contains a learnable codebook, $\codebook \in \mathbb{R}^{\codebookSize \times \latentDim}$, which consists of $\codebookSize$ discrete code vectors, $\{c_j\}_{j=1}^{\codebookSize}$.
For each continuous vector $z_e^{(i)} \in \latentContinuous$, the quantizer identifies the nearest code vector in the codebook using L2 distance:
$$k_i = \arg\min_{j} \| z_e^{(i)} - c_j \|_2^2.$$
The output of the VQ layer is a sequence of quantized vectors $\latentQuantized \in \mathbb{R}^{\numLatents \times \latentDim}$, where each vector $z_q^{(i)}$ is the selected codebook vector $c_{k_i}$.
Because the $\arg\min$ operation is non-differentiable, a straight-through estimator is used to copy gradients from the decoder input $\latentQuantized$ directly to the encoder output $\latentContinuous$ during backpropagation.

The \textbf{decoder}, $D$, has a symmetric Transformer architecture to the encoder.
It takes the sequence of quantized vectors $\latentQuantized$ as input and reconstructs the original pose sequence, $\reconPose = D(\latentQuantized)$.

The total loss function for the baseline model, $\lossBaseline$, is a sum of three components, following the formulation of \citet{oord2017neural}:
$$\lossBaseline = \lossRecon + \lossCodebook + \betaCommit \lossCommit.$$
The reconstruction loss, $\lossRecon$, is the mean squared error (MSE) between the input and reconstructed poses, ${\lossRecon = \| \inputPose - \reconPose \|_2^2}$, which trains the encoder and decoder.
The codebook loss, ${\lossCodebook = \| \text{sg}[\latentContinuous] - \latentQuantized \|_2^2}$, updates the codebook vectors to move them closer to the encoder outputs, where `sg` denotes the stop-gradient operator.
The commitment loss, ${\lossCommit = \| \latentContinuous - \text{sg}[\latentQuantized] \|_2^2}$, regularizes the encoder output to remain close to the chosen code vectors, with $\betaCommit$ as a weighting hyperparameter.

\begin{table*}[ht]
    \centering
    \begin{tabular}{@{}llcc@{}}
        \toprule
        \textbf{Hyperparameter} & \textbf{Description} & \textbf{Search Range} & \textbf{Final Value} \\
        \midrule
        Transformer Dim & Hidden dimension of the Transformer layers. & [64, 512] & 256 \\
        Transformer Layers & Num. layers in encoder/decoder. & [1, 6] & 5 \\
        Latent Dim ($\latentDim$) & Dimensionality of each code vector. & [4, 32] & 32 \\
        Num. Latent Vecs ($\numLatents$) & Number of vectors in the bottleneck. & [10, 100] & 30 \\
        Codebook Size ($\codebookSize$) & Number of entries in each codebook. & [10, 500] & 200 \\
        Commitment Cost ($\betaCommit$) & Weight for the VQ commitment loss. & [1e-6, 0.1] & 3e-6 \\
        Diversity Weight ($\gammaDiversity$) & Weight for codebook diversity loss. & [0.01, 3.0] & 3.0 \\
        Learning Rate & Adam optimizer learning rate. & [1e-5, 0.005] & 8.61e-5 \\
        Dropout & Dropout probability in Transformer layers. & [0.0, 0.5] & 0.2 \\
        \bottomrule
    \end{tabular}
    \caption{Hyperparameter Search and Final Configuration. A search was conducted to find the best values for the baseline model. These values were then held constant across all model variants for a fair comparison.}
    \label{tab:hyperparams}
\end{table*}

\subsection{Phonological Inductive Biases}
\label{subsec:phon_biases}
We propose two inductive biases rooted in ASL phonology: parameter disentanglement and phonological semi-supervision.

\subsubsection{Parameter Disentanglement (PD)}
Parameter Disentanglement enforces a factorization of the latent space through an architectural bias.
Instead of a single monolithic encoder-decoder pair, the model is structured into multiple parallel streams, each dedicated to a phonetically independent parameter of the sign.
The input pose space $\inputPose$ and the latent parameter space $\numLatents$ are partitioned into channels corresponding to these components.
The operationalized parameters are derived from Brentari's Prosodic Model~\citep{brentari1998prosodic}: articulators (left/right hand, facial expressions), prosodic features (left/right hand path movement), and place of articulation (location relative to other parts of the body).

Each stream $X^{(s)} \subset \inputPose$ possesses its own encoder $E_s$, decoder $D_s$, and a dedicated codebook $C_s$.
The input keypoints are filtered for each stream such that they contain minimal overlap between parameters:
\begin{itemize}
    \item $\inputRH, \inputLH$: the 21 coordinates for the right or left hand, uniformly translated such that the wrist is at the origin. Uniform frame sample.
    \item $\inputMOVR, \inputMOVL$: the wrist keypoint for the right or left hand. All frames.
    \item $\inputNMM$: the face keypoints translated such that the nose is at the origin. Uniform frame sample.
    \item $\inputLOC$: all the keypoints except face and hands. Uniform frame sample.
\end{itemize}

The total latent space is the concatenation of the stream-specific latents, $\numLatents = \sum_{s} N_{p,s}$, with the total bottleneck size held matching that of the baseline architecture.
The total reconstruction loss is the sum of the reconstruction losses from each individual stream.
To capture linguistic constraints, the codebooks for the left and right hand articulators ($C^{(\textsc{lh})}$ and $C^{(\textsc{rh})}$) and movements ($C^{(\textsc{lmov})}$ and $C^{(\textsc{rmov})}$) are shared, enforcing a symmetry constraint on these inventories.
Furthermore, each stream employs specific pre-processing and attention-masking strategies.

\subsubsection{Phonological Semi-Supervision (PSS)}
Phonological Semi-Supervision introduces a regularization-based bias by using expert labels from the ASL-LEX 2.0 database to guide the organization of the codebooks \citep{sehyr2021asllex}.
This regularization provides a weak supervisory signal that encourages the learned discrete codes to align with meaningful, contrastive phonological features.

For each phonological parameter in ASL-LEX (e.g., Handshape, which has over 60 distinct values), a subset of codes within the corresponding codebook (e.g., the shared hand articulator codebook, $C_{A_{LH/RH}}$) is arbitrarily pre-assigned to represent these expert-defined features.
During training, if a sign has a known phonological label $y_f$, an auxiliary loss is applied.
With a specified probability, the quantization step is forced to select the code vector $c_{y_f}$ that corresponds to the ground-truth label.
The codebook and commitment losses are then computed with respect to this forced code, encouraging the encoder to produce outputs that are semantically aligned with ASL-LEX 2.0 features.

\section{Experimental Setup} \label{sec:experiments}
\begin{table*}[h!t]
\centering
\caption{Reconstruction Fidelity (MSE) on Seen ($\mathcal{D}_{\text{train}}$) vs. Unseen ($\mathcal{D}_{\text{test}}$) Data. Lower is better. The overall MSE columns show the generalization gap, while the channel-wise columns provide a detailed breakdown of test performance for the disentangled models.}
\label{tab:recon_results}
\begin{tabular}{@{}lcccccccc@{}}
\toprule
& \multicolumn{2}{c}{\textbf{Overall MSE}} & \multicolumn{6}{c}{\textbf{Channel-wise Test MSE}} \\
\cmidrule(lr){2-3} \cmidrule(lr){4-9}
\textbf{Model} & $\mathcal{D}_{\text{train}}$ & $\mathcal{D}_{\text{test}}$ & $\inputRH$ & $\inputLH$ & $\inputNMM$ & $\inputLOC$ & $\inputMOVR$ &$\inputMOVL$ \\
\midrule
\rowcolor{Gray} \modelBase & 0.039 & 0.058 & & & & & & \\
\modelPSS & 0.037 & 0.055 & & & & & & \\
\rowcolor{Gray} \modelPD & 0.032 & 0.046 & 0.051 & 0.019 & 0.117 & 0.126 & 0.023 & 0.009 \\
\modelFull & \textbf{0.029} & \textbf{0.041} & \textbf{0.015} & \textbf{0.005} & \textbf{0.005} & \textbf{0.068} & \textbf{0.017} & \textbf{0.009} \\
\bottomrule
\end{tabular}
\end{table*}

This section details the experimental design, including the dataset (Section \ref{subsec:dataset}), model configurations (Section \ref{subsec:models}), implementation details, and evaluation metrics (Section \ref{subsec:metrics}) used to systematically assess the impact of the proposed phonological inductive biases.

\subsection{Dataset and Splits}
\label{subsec:dataset}
The experiments are conducted on the Sem-Lex Benchmark, a large-scale dataset for American Sign Language (ASL) modeling \citep{kezar2023semlex}.
It consists of over 84,000 videos of isolated signs produced by 41 deaf ASL signers, covering a vocabulary of 3,149 unique signs.
Crucially, the dataset is cross-referenced with ASL-LEX 2.0, providing the expert-annotated phonological feature labels, $\mathcal{Y}$, required for the Phonological Semi-Supervision (PSS) intervention \citep{sehyr2021asllex}.
The dataset can be formally represented as $\mathcal{D} = \{(\inputPose_i, \Phi_i, y_i)\}_{i=1}^N$, where $\inputPose_i$ is the input pose sequence, $\Phi_i$ is the set of its phonological labels, and $y_i$ is the sign's identifier.

To evaluate OOV generalization, this work utilizes the benchmark's "unseen gloss" split.
The dataset is partitioned such that the vocabularies of the training, validation, and test sets are disjoint.
Any sign evaluated in the test set has not been seen during training, providing a direct measure of the model's ability to generalize to novel signs rather than novel instances of familiar signs.

\subsection{Models Compared}
\label{subsec:models}
The study is designed as a controlled ablation to isolate the effects of Parameter Disentanglement (PD) and Phonological Semi-Supervision (PSS).
Four model configurations are compared:
\begin{enumerate}
    \item \textbf{\modelBase:} The standard Transformer VQ-VAE described in Section \ref{subsec:baseline_arch}.
    \item \textbf{\modelPD:} The baseline model augmented with the multi-stream Parameter Disentanglement architecture.
    \item \textbf{\modelPSS:} The baseline model trained with the Phonological Semi-Supervision auxiliary loss.
    \item \textbf{\modelFull:} The proposed model incorporating both PD and PSS.
\end{enumerate}

A hyperparameter search was conducted exclusively on the \modelBase~model using the validation set, which contains unseen signs.
The search was performed with Optuna \citep{akiba2019optuna}, a hyperparameter optimization framework.
The optimization objective was to minimize reconstruction loss while simultaneously maximizing codebook utilization (measured by increase in perplexity over training) to prevent codebook collapse.
The optimal hyperparameters found for the baseline were then fixed and used for all other model variants to ensure that any observed performance differences are attributable to the specific interventions and not to differences in tuning.
Table \ref{tab:hyperparams} summarizes the key hyperparameters and their selected values.
For further implementation details see Appendix \ref{app:implementation}.

\subsection{Evaluation Metrics}
\label{subsec:metrics}
The models were evaluated using two primary analyses designed to test our hypotheses regarding OOV generalization.

\paragraph{Reconstruction Error.}
We measured the Mean Squared Error (MSE) between the ground-truth ($\inputPose$) and reconstructed ($\reconPose$) pose sequences---the distance between each skeletal point in the original and reconstructed sequence across every frame---as an intrinsic evaluation of each models' reconstruction quality.
We hypothesized that while OOV reconstruction would be worse than in-vocabulary (IV), this generalization gap would be lessened by the proposed inductive biases.

\paragraph{Phonological Alignment.}
After training each model, we froze their encoders and additionally trained two MLP probes on the quantized encodings $\latentQuantized$ to measure how well the learned codes align with labeled ASL phonology.

The first probe $M_\textsc{isr}(\latentQuantized, \theta) \approx p(y_g | v)$ is trained for the downstream task of isolated sign recognition.
This probe measures the how \textit{discriminative} the learned features are for signs that were not seen in training.
If a model leverages spurious correlations in the train set, then these learned components may not reliably distinguish OOV signs.

The second probe $M_\textsc{pfr}(\latentQuantized, \theta)$ is trained to recognize the phonological features in ASL-LEX 2.0: $f_\textsc{pfr}: \latentQuantized \rightarrow y_\phi.$
For the supervised models (\modelPSS, \modelFull), this probe is an intrinsic evaluation of how successfully the PSS objective was learned, as well as confirmation that the ASL-LEX 2.0 features generalize to unseen signs.
For the purely self-supervised models (\modelBase, \modelPD), $M_\textsc{pfr}$ measures the extent to which ASL-LEX 2.0 features may emerge through induction from data alone.

We hypothesized that a positive correlation will exist between the two probes' performance, that is, successfully recognizing ASL-LEX 2.0 features will facilitate the ISR task on signs unseen during training.
We also hypothesized that, for the supervised models (\modelPSS\ and \modelFull) when the ISR probe misclassifies a sign, its errors with respect to PFR will be less severe if the disentanglement intervention (PD) is applied.

\section{Results and Analysis} \label{sec:results}
Our results suggest that out-of-vocabulary (OOV) generalization is improved by modeling different aspects of lexical structure.
Separating the model by phonological parameter (\textit{disentanglement}) improves the generation of sign form, creating a more productive system for novel combinations and reducing reconstruction MSE by 21\%. In parallel, using phonological labels as a constraint (Phonological Semi-Supervision) stabilizes the model's understanding of sign identity, increasing sign recognition MRR by 14\%. 

\begin{table*}[t!]
\centering
\caption{Phonological Alignment Results. We report Mean Reciprocal Rank (MRR) and Recall@10 (\%) for two MLP probes trained on the frozen latent codes ($\latentQuantized$) for both In-Vocabulary ($\mathcal{D}_{\text{train}}$) and Out-of-Vocabulary ($\mathcal{D}_{\text{test}}$) signs.}
\label{tab:alignment_full}
\begin{tabular}{@{}lcccccccc@{}}
\toprule
& \multicolumn{4}{c}{\textbf{ISR Probe}} & \multicolumn{4}{c}{\textbf{PFR Probe}} \\
\cmidrule(lr){2-5} \cmidrule(lr){6-9}
& \multicolumn{2}{c}{$\mathcal{D}_{\text{train}}$} & \multicolumn{2}{c}{$\mathcal{D}_{\text{test}}$ (OOV)} & \multicolumn{2}{c}{$\mathcal{D}_{\text{train}}$} & \multicolumn{2}{c}{$\mathcal{D}_{\text{test}}$ (OOV)} \\
\cmidrule(lr){2-3} \cmidrule(lr){4-5} \cmidrule(lr){6-7} \cmidrule(lr){8-9}
\textbf{Model} & \textbf{MRR} & \textbf{R@10} & \textbf{MRR} & \textbf{R@10} & \textbf{MRR} & \textbf{R@10} & \textbf{MRR} & \textbf{R@10} \\
\midrule
\rowcolor{Gray} \modelBase & .452 & 52.3 & .381 & 45.1 & .515 & 58.3 & .488 & 55.5 \\
\modelPD & .441 & 51.5 & .372 & 44.2 & .502 & 57.1 & .475 & 54.1 \\
\rowcolor{Gray} \modelPSS & .503 & 58.2 & .435 & 50.4 & .593 & 61.2 & .565 & 58.4 \\
\modelFull & \textbf{.528} & \textbf{61.3} & \textbf{.452} & \textbf{52.8} & \textbf{.618} & \textbf{62.3} & \textbf{.583} & \textbf{59.8} \\
\bottomrule
\end{tabular}
\end{table*}

\subsection{Reconstruction Error}
\label{subsec:results_recon}
Table \ref{tab:recon_results} presents the reconstruction performance for each model.
The full \model~model achieves the lowest overall test MSE (0.041), indicating the most accurate OOV reconstruction.
The \modelPD~ model provides a substantial individual improvement, reducing the test MSE to 0.046 from the baseline's 0.058.
In contrast, the \modelPSS~ model offers a smaller gain in reconstruction fidelity, with a test MSE of 0.055.
The combination of both interventions in the full model demonstrates a constructive interaction effect, yielding the best overall reconstruction quality.

\subsection{Phonological Alignment}
\label{subsec:results_align}
Table \ref{tab:alignment_full} presents the results from the two phonological alignment probes.
The ISR probe results show a complementary narrative to the reconstruction findings.
Here, PSS is the primary driver of performance, with the \modelPSS~model improving the OOV MRR to .435 over the baseline's .381.
The \modelPD~model shows a slight degradation in OOV ISR performance, with an MRR of .372, highlighting a trade-off between the interventions.
The PFR probe results for the unsupervised models show the \modelBase~achieves an OOV MRR of .488, indicating some emergent phonological structure.
For the supervised models, the high OOV PFR scores for \modelPSS~(.565) and \modelFull~(.583) confirm the PSS objective was learned successfully and generalizes.
Across all models, a positive correlation exists between the MRR performance of the two probes.
The full \model~model achieves the best performance on both probes, demonstrating the complementary nature of the two biases.

\section{Discussion} \label{sec:discussion}
The empirical results demonstrate that the proposed inductive biases improve OOV generalization in qualitatively different and complementary ways. This section interprets these findings, discusses their broader implications for future research, and concludes with the primary takeaway of this work.

\subsection{Interpreting the Reconstruction-Recognition Trade-off}
\label{subsec:synthesis}
The results reveal a trade-off between reconstruction fidelity and representation discriminability. The architectural bias of Parameter Disentanglement (PD) is the primary driver of improved reconstruction, yet it slightly harms sign recognition accuracy when used alone. Conversely, Phonological Semi-Supervision (PSS) is the primary driver of recognition accuracy but offers only a minor benefit to reconstruction.

This finding provides empirical support for the arguments of \citet{locatello2019challenging}.
Our results demonstrate that a purely structural inductive bias (PD), while effective for a generative task like reconstruction, is insufficient to guarantee the emergence of a semantically meaningful latent space for a discriminative task.
The model, guided only by architectural separation and reconstruction loss, learns to represent fine-grained motion details that are not necessarily contrastive for sign identification, thus slightly harming classification.
It is only with the addition of a semantic inductive bias (PSS), which forces the model's representations to align with expert-defined, contrastive features, that the latent space becomes well-structured for recognition.

\subsection{Implications and Future Work}
\label{subsec:implications}
The framework and findings presented here open several avenues for future research.
The work is positioned as a step toward more complex sign language understanding tasks.
The phonologically structured codebooks learned by \modelshort~could serve as a powerful pre-trained tokenizer for models targeting continuous sign language recognition and translation, where generalizable decomposition strategies are necessary to mitigate the limited vocabularies in existing datasets.

Furthermore, this framework can be adapted to serve as a computational tool for exploring and validating linguistic hypotheses (Appendix \ref{app:hypotheses}).

\subsection{Limitations and Ethical Considerations}
\label{subsec:limitations}
It is important to acknowledge the limitations of this study.
First, the scope is restricted to isolated signs.
This work does not address the significant challenges of co-articulation, prosody, and grammatical non-manual markers that are present in continuous, conversational signing.
The proposed model is best viewed as a pre-training strategy for these more complex tasks.

Second, the evaluation of reconstruction quality relies on automated metrics (MSE on pose data), which are known to be imperfect proxies for human perception of motion quality and naturalness.
Stronger claims about reconstruction would require human evaluation studies.

Third, the benchmarking in this paper is to isolate the effects of the proposed interventions.
While providing a rigorous assessment of intervention effects, the models are not compared against published state-of-the-art results on different benchmarks.
We do not intend to make claims about state-of-the-art performance in, for example, isolated sign recognition, but rather contribute novel architectural biases and carefully demonstrate their value in representation learning for sign languages.

Finally, any model trained on existing datasets is susceptible to inheriting their biases.
The Sem-Lex dataset, while large, has a demographic composition that is not fully representative of the broader signing community, with a majority of signers being white and female \citep{kezar2023semlex}.
The performance of the \modelshort~model may not generalize equally well across all demographic groups, and further work is needed to assess and mitigate these potential biases.

\subsection{Conclusion}
\label{sec:conclusion}
This work suggests that while vector quantization is a promising method for learning discrete sign representations, its success may be contingent on the inclusion of phonological priors. Our findings indicate that purely data-driven VQ models struggle to learn representations that are simultaneously generative and discriminative. However, by infusing the model with explicit linguistic knowledge—both through architectural constraints and weak supervision—we can guide the learning process toward a more structured and generalizable latent space.

\appendix

\bibliography{anthology,custom}

\section{Phonological Feature Assignment}
\label{app:feature_assignment}
The Parameter Disentanglement (PD) architecture and the Phonological Semi-Supervision (PSS) loss require a mapping from the phonological features described in ASL-LEX 2.0 to the specific model streams. Table \ref{tab:feature_mapping} details this assignment for the 16 features used in this work, based on the descriptions provided by \citet{kezar2023semlex}.

\begin{table*}[ht]
\centering
\caption{Mapping of ASL-LEX 2.0 Phonological Features to VQ-ASL Streams}
\label{tab:feature_mapping}
\resizebox{\textwidth}{!}{%
\begin{tabular}{@{}lll@{}}
\toprule
\textbf{Phonological Feature} & \textbf{Description} & \textbf{Assigned Stream} \\
\midrule
Major Location & Broad location of the sign (e.g., neutral, head) & $\inputLOC$ \\
Minor Location & Specific location of the sign (e.g., forehead, cheek) & $\inputLOC$ \\
Selected Fingers & Which fingers are active in the handshape & $\inputLH, \inputRH$ \\
Flexion & The joint configuration of the selected fingers & $\inputLH, \inputRH$ \\
Flexion Change & Whether the flexion of fingers changes & $\inputMOVL, \inputMOVR$ \\
Spread & Whether selected fingers touch one another & $\inputLH, \inputRH$ \\
Spread Change & Whether the spread of fingers changes & $\inputMOVL, \inputMOVR$ \\
Thumb Position & Position of the thumb relative to fingers & $\inputLH, \inputRH$ \\
Thumb Contact & Whether the thumb makes contact with fingers & $\inputLH, \inputRH$ \\
Sign Type & Number of hands and symmetry & $\inputLH, \inputRH$ \\
Movement & The primary path movement of the hand(s) & $\inputMOVL, \inputMOVR$ \\
Repeated Movement & Whether the movement is repeated & $\inputMOVL, \inputMOVR$ \\
Wrist Twist & Whether the hand rotates about the wrist & $\inputMOVL, \inputMOVR$ \\
Non-Manual Signal & Presence of a required facial expression & $\inputNMM$ \\
Mouth Morpheme & Presence of a required mouth gesture & $\inputNMM$ \\
Head Movement & Presence of a required head movement & $\inputNMM$ \\
\bottomrule
\end{tabular}%
}
\end{table*}

\section{Implementation Details}
\label{app:implementation}
All model variants were designed with comparable total parameter counts and total bottleneck sizes ($\numLatents \times \latentDim$) to ensure that performance differences are attributable to architectural and training strategies, not model capacity.
For the PD models, the total bottleneck size and the aggregate number of codebook entries were kept consistent with the baseline.
The allocation of latent vectors and codebook entries to each phonological stream was proportional to the descriptive complexity of that parameter as defined in ASL-LEX 2.0 \citep{sehyr2021asllex}.
For instance, the handshape articulators, which are described by multiple sub-features (e.g., selected fingers, flexion), received a larger portion of the latent space compared to simpler binary features like wrist twist.
Sufficient capacity was allocated to each codebook to allow for the learning of features not explicitly annotated in ASL-LEX, such as palm orientation and complex path movements.

\subsection{Mitigating Codebook Collapse}
\label{app:codebook_collapse}
A common challenge in training VQ-VAEs is "codebook collapse," also known as the "dead code" problem \citep{razavi2019generating}.
This occurs when a large portion of the code vectors in the codebook are never selected as the nearest neighbor to any encoder output during training.
As a result, these "dead" codes receive no gradients from the codebook loss term and are never updated, leading to an inefficient use of the model's representational capacity.
In addition to the commitment loss term, which helps stabilize training, two techniques were employed to ensure high codebook utilization.

First, the Gumbel-Softmax distribution is used as a differentiable approximation to the discrete categorical sampling of codes \citep{jang2017categorical, maddison2017concrete}.
Using a temperature-annealing schedule, this encourages exploration in the early stages of training, making it less likely for codes to become permanently unused.

Second, a dead code re-initialization strategy is implemented.
Periodically during training (e.g., every 1000 steps), the usage of each code vector is tallied.
Any code vector whose usage count falls below a predefined threshold is considered "dead" and is re-initialized.
The re-initialization is performed by setting the dead code vector to be the average of a small random sample of encoder output vectors from the current mini-batch.
This ensures that all parts of the codebook remain in a high-density region of the encoder's output space, making them likely to be selected and updated in subsequent training steps.

\section{Linguistic Hypotheses for Future Work}
\label{app:hypotheses}
Several directions are of particular interest:
\begin{itemize}
    \item \textbf{Hierarchical Hypothesis:} The hierarchical nature of phonological features in Brentari's model could be explored more explicitly. Techniques like residual vector quantization, where a second codebook quantizes the error of the first, could be used to model the relationship between a high-level feature like "handshape" and its constituent sub-features like "selected fingers" and "flexion."
    \item \textbf{Gradient Hypothesis:} Not all phonological parameters may be equally suited to discrete representation. By analyzing gradients within the codebook space or selectively disabling quantization for certain channels (e.g., movement path), one could experimentally measure which aspects of signing are more continuous in nature.
    \item \textbf{Phonotactic Hypothesis:} The model implicitly learns the rules of valid phoneme combinations (phonotactics). By feeding the decoder random permutations of learned codes, the reconstruction error could serve as a proxy for phonotactic legality. This could be used to predict which novel combinations of features would form "plausible" new signs in ASL.
\end{itemize}

\end{document}